\documentclass[sigconf,table, authorversion, nonacm]{acmart}

\AtBeginDocument{%
  \providecommand\BibTeX{{%
    \normalfont B\kern-0.5em{\scshape i\kern-0.25em b}\kern-0.8em\TeX}}}

\usepackage[utf8]{inputenc}
\usepackage[T1]{fontenc}
\usepackage{dirtytalk}
\MakeRobust{\say} %
\usepackage{multirow}
\usepackage{amsmath}

\usepackage{pgfplots}%
\usepackage{pgf}
\usepackage[capitalize]{cleveref} %
\usepackage{xspace} %
\usepackage{pifont} %
\usepackage{microtype} %
\usepackage{subcaption} %
\usepackage{makecell} %
\usepackage[inline]{enumitem}
\usepackage{collcell} %
\usepackage{hhline}
\usepackage{siunitx}
\usepackage{etoolbox}

\makeatletter
\def\@copyrightspace{\relax}
\makeatother

\makeatletter
\DeclareRobustCommand\onedot{\futurelet\@let@token\@onedot}
\def\@onedot{\ifx\@let@token.\else.\null\fi\xspace}
\newcommand{\etal}{\emph{et~al\onedot}}
\newcommand{\ie}{i.\,e.,\xspace}
\newcommand{\eg}{e.\,g.,\xspace}

\newcommand{\cf}{cf\onedot}

\begin{document}

\title{SniffyArt: The Dataset of Smelling Persons}

\author{Mathias Zinnen}
\affiliation{%
\institution{Pattern Recognition Lab, Friedrich-Alexander-Universität}
\city{Erlangen}
\country{Germany}
\orcid{0000-0003-4366-5216}
}
\author{Azhar Hussian}
\orcid{0009-0008-8125-7081}
\affiliation{%
\institution{Pattern Recognition Lab, Friedrich-Alexander-Universität}
\city{Erlangen}
\country{Germany}
}
\author{Hang Tran}
\orcid{0009-0004-0812-9316}
\affiliation{
\institution{Pattern Recognition Lab, Friedrich-Alexander-Universität}
\city{Erlangen}
\country{Germany}
}
\author{Prathmesh Madhu}
\affiliation{%
\institution{Pattern Recognition Lab, Friedrich-Alexander-Universität}
\city{Erlangen}
\country{Germany}
\orcid{0000-0003-2707-415X}
}
\author{Andreas Maier}
\affiliation{%
\institution{Pattern Recognition Lab, Friedrich-Alexander-Universität}
\city{Erlangen}
\country{Germany}
\orcid{0000-0002-9550-5284}
}
\author{Vincent Christlein}
\affiliation{%
\institution{Pattern Recognition Lab, Friedrich-Alexander-Universität}
\city{Erlangen}
\country{Germany}
\orcid{0000-0003-0455-3799}
}

\renewcommand{\shortauthors}{Mathias Zinnen et al.}

\begin{abstract}
Smell gestures play a crucial role in the investigation of past smells in the visual arts yet their automated recognition poses significant challenges.
This paper introduces the SniffyArt dataset, consisting of 1941 individuals represented in 441 historical artworks. 
Each person is annotated with a tightly fitting bounding box, 17 pose keypoints, and a gesture label. 
By integrating these annotations, the dataset enables the development of hybrid classification approaches for smell gesture recognition. 
The dataset's high-quality human pose estimation keypoints are achieved through the merging of five separate sets of keypoint annotations per person. 
The paper also presents a baseline analysis, evaluating the performance of representative algorithms for detection, keypoint estimation, and classification tasks, showcasing the potential of combining keypoint estimation with smell gesture classification. 
The SniffyArt dataset lays a solid foundation for future research and the exploration of multi-task approaches leveraging pose keypoints and person boxes to advance human gesture and olfactory dimension analysis in historical artworks.
\end{abstract}

\begin{teaserfigure}
  \includegraphics[width=\textwidth]{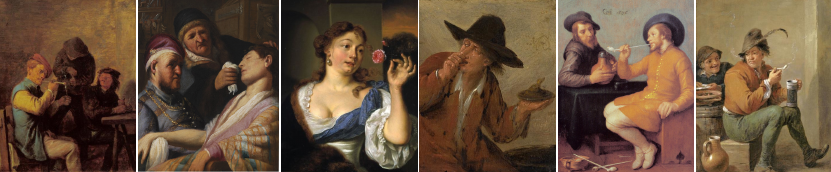}
  \caption{Samples from the dataset displaying various smell gestures.\protect\footnotemark}
  \Description{Six samples from the SniffyArt dataset, displaying a variety of smell gestures.}
  \label{fig:teaser}
\end{teaserfigure}

\maketitle

{
\renewcommand{\thefootnote}{\fnsymbol{footnote}}
\footnotetext[1]{Image credits (left to right, all cropped):
  \textit{Three peasants somking in an interior}. Adriaen Brouwer. c.1624--1625. RKD -- Netherlands Institute for Art History, RKDimages (241587). Public Domain.
  \textit{Unconscious Patient}. Rembrandt van Rijn. 1620--1638. RKD -- Netherlands Institute for Art History, RKDimages (283757). Public Domain.
  \textit{Young woman with a rose}. Ary de Vois. 1653--1680. RKD -- Netherlands Institute for Art History, RKDimages (18766). Public Domain.
  \textit{The five senses: smell}. Jan Molenaer (II). 1670--1700. RKD -- Netherlands Institute for Art History, RKDimages (278370). Public Domain.
  \textit{Twee mannen met bierpul en pijp}. Cornelis Cornelisz. van Haarlem. 1636. RKD -- Netherlands Institute for Art History, RKDimages (55038). Public Domain.
  \textit{A peasant smoking and an old woman}. David Teniers (II). 1651--1690. RKD -- Netherlands Institute for Art History, RKDimages (4458). Public Domain.
  } 
}

\section{Introduction}
Smells play a crucial role in shaping human everyday experience, influencing emotions, memories and behaviour.
Despite their ubiquitousness, they rarely cross the threshold of our consciousness. %
Recently, the significance of smell has increasingly been acknowledged in the field of cultural heritage~\cite{verbeek2013inhaling,bembibre2017smell} and the humanities~\cite{tullett2021state, leemans2022whiffstory,jenner2011follow}.
Specifically in digital heritage and computational humanities, the role of smells is gaining more and more prominence~\cite{van2023more,lisena2022capturing,menini-etal-2023-scent}.
Tracing past smells and their societal roles can be achieved through the identification of olfactory references in artworks and visual media. 
However, the inherent invisibility of smells poses a significant challenge in this endeavour. 
Recognising olfactory references requires the detection of proxies such as smell-active objects, fragrant spaces, or olfactory iconography which indirectly indicate the presence of smells~\cite{zinnen2021see}.
Among these proxies, smell gestures, such as reactions to smell or smell-producing actions, provide the most explicit gateway to the olfactory dimensions of a painting.
However, recognizing smell gestures is a particularly challenging task, as they exhibit high intra-class variance, are difficult to precisely localize, and their identification involves a higher degree of subjectivity. 
As a first step towards recognizing smell gestures, we present the SniffyArt dataset, annotated with person boxes, pose estimation keypoints, and smell gesture labels.
By combining these three types of annotations, we aim to facilitate the development of novel gesture recognition methods that leverage all three label types. 
Furthermore, we evaluate various baseline approaches for person detection, keypoint estimation, and smell gesture classification using this dataset.
Our contributions are as follows:
\begin{itemize}
    \item We introduce the SniffyArt dataset, featuring artworks annotated with bounding boxes, pose estimation keypoints, and smell gesture annotations for nearly 2000 persons.
    \item We evaluate initial baseline methods for person detection, keypoint estimation, and smell gesture recognition on the SniffyArt dataset.
\end{itemize}
Through this work, we hope to advance research in the domain of smell gesture classification and pave the way for a deeper understanding of olfactory dimensions in visual art and cultural heritage.

\section{Related Work}
\paragraph{Computer Vision and the Humanities}
Many computer vision tasks like object detection, human pose estimation, or image segmentation have had their main research focus on real-world images.
The availability of large-scale photographic datasets like ImageNet~\cite{russakovsky2015imagenet}, COCO~\cite{lin2014microsoft}, OpenImages~\cite{kuznetsova2020open}, or Objects365~\cite{shao2019objects365} has enabled computer vision methods to achieve impressive performance on natural images.
Applying those methods to digital humanities and cultural heritage can provide a valuable addition to traditional methods of the humanities~\cite{bell2016visuelle, bell2018computer}.
It enables humanities scholars to complement their analysis with a data-driven perspective, thus broadening their view and enabling them to perform \say{distant viewing}~\cite{arnold2019distant}.
Unfortunately, when applying standard architectures on artwork images, we observe a significant performance drop, which has been attributed to the domain shift problem~\cite{cai2015cross,hall2015cross,cetinic2022understanding}.
This domain mismatch can be tackled by applying domain adaptation techniques~\cite{farahani2021brief} to overcome the representational gap between artworks and real-world images.
Various researchers have proposed the application of style transfer~\cite{kadish2021improving,lu2022data,madhu2022enhancing}, transfer learning~\cite{sabatelli2018deep, gonthier2021analysis, zinnen2022transfer,zhao2022big}, or the combination of multiple modalities~\cite{radford2021learning,ali2022musti,kiymet2022multimodal,garcia2018read,gupta2020towards}.

\paragraph{Person Detection}
The task to detect persons can be considered a special case of the more generic object detection task. 
Object detection algorithms are usually categorised as one-stage, two-stage, and more recently transformer-based approaches~\cite{jiao2019survey}.
Two-stage algorithms propose candidate regions of interest in the first step and refine and classify those regions in the second step.
The most prominent two-stage algorithms are representatives of the R-CNN~\cite{girshick2014rich} family. 
With various tweaks and refinements~\cite{girshick2015fast, ren2015faster, lin2017feature,cai2018cascade}, R-CNN-based algorithms still provide competitive results today. 
Due to its canonical role, we will apply the R-CNN based detector Faster R-CNN~\cite{ren2015faster} to generate baseline results for our experiments.
One-stage algorithms, on the other hand, merge the two stages and operate on a predefined grid. 
On this grid, candidate objects are simultaneously predicted and classified in a single step, thus achieving a higher inference speed.
The best-known examples of one-stage algorithms are You Only Look Once (YOLO)~\cite{redmon2016you} architecture and descendants~\cite{redmon2017yolo9000, redmon2018yolov3,jocher2020ultralytics,terven2023comprehensive}.
In contrast to these paradigms, our approach is based on transformer detection heads as proposed by Carion \etal~\cite{carion2020end} in their Detection Transfomer (DETR) architecture.
In DETR and derivatives~\cite{zhu2020deformable, zhang2022dino} a set of predicted candidate boxes are assigned to ground truth boxes by solving a set assignment problem using the Hungarian Algorithm~\cite{kuhn1955hungarian}.
DETR-based algorithms, most notably DINO~\cite{zhang2022dino}, set the current state of the art in object detection in natural images.
DETR-based algorithms, most notably DINO~\cite{zhang2022dino}, set the current state of the art in object detection in natural images.

In the artistic domain, pioneering work by \cite{crowley2014state, crowley2015search, crowley2016art} has opened the field of object recognition in the visual arts. 
Gonthier \etal ~\cite{gonthier2018weakly} proposed a weakly supervised approach to cope with the shortage of object-level labels in artworks and published the IconArt dataset consisting of about 5000 instances within 10 iconography-related classes~\cite{gonthier_nicolas_2018_4737435}.
Going in the same direction, Madhu \etal~\cite{madhu2022one} propose a one-shot algorithm that enables the detection of unseen objects in artworks. 
Specifically for person detection, Westlake \etal~\cite{westlake2016detecting} provide the PeopleArt dataset and evaluate a Fast-RCNN on the dataset.
In the ODOR challenge~\cite{zinnen2022odor}, participants were given the task of detecting a set of 87 smell-related objects depicted in historical artworks. 
The recent introduction of the DeART dataset~\cite{reshetnikov2022deart} promises to advance the field further by providing more than 15,000 artworks annotated with object-level annotations across 70 categories.

\paragraph{Human Pose Estimation (HPE)}
The estimation of body poses is achieved via the regression of a set of keypoints corresponding to body joints that define a person's pose.
In practice, many modern pose estimation algorithms do not directly regress the exact keypoints but operate on heatmaps indicating the probability distribution for keypoint existence in a region. 
The set of keypoints defining the body pose can be defined in multiple ways. 
In this work, we consider the definition of body joints defined by Lin \etal~\cite{lin2014microsoft}.
Pose estimation algorithms can be roughly grouped into bottom-up or top-down approaches~\cite{zheng2020deep}.
In bottom-up algorithms~\cite{cao2017realtime, kreiss2019pifpaf, cheng2020higherhrnet, geng2021bottom}, keypoints are detected first and assigned to specific persons afterwards whereas top-down algorithms~\cite{cai2020learning,cai2018cascade,sun2019deep,xu2022vitpose,xiao2018simple} require a person-detection stage before estimating the pose keypoints.
Recent state-of-the-art pose estimation networks combine a two-stage pipeline with transformer-based keypoint regression heads.
Zhang \etal~\cite{zhang2021towards} demonstrated that an additional skeleton refinement step can further increase the estimation accuracy.

Applications in the artistic domain suffer from a lack of large-scale annotated datasets, which is even worse than in the case of object detection.
Springstein \etal~\cite{springstein2022semi} tackle this lack of annotated data by training on stylised versions of the COCO dataset and applying the semi-supervised soft-teacher approach~\cite{xu2021end}.
A recent application of HPE for artwork analysis has been presented by Zhao \etal~\cite{zhao2022automatic} who combine body segmentation, HPE, and hierarchical clustering to analyze body poses in a dataset of c. 100k artworks. 

Apart from the yet unpublished PoPArt~\cite{schneider2023poses} dataset, the proposed SniffyArt dataset constitutes the first artwork dataset with keypoint-level annotations.

\section{SniffyArt Dataset}
\subsection{Data Collection}
\label{sec:coll}
The data was collected and annotated in three phases: preselection, person annotation, and keypoint annotation.

In the \textit{preselection} phase, we automatically annotated a large set of candidate artworks from various digital museum collections with 139 smell-active objects. 
From this annotated set of artworks, we selected about 2000 images containing depictions of smell gestures.
The object annotations served as cues to facilitate the search; \eg by filtering for images containing pipes when looking for ``smoking'' gestures.
Filtering and tagging in this phase was achieved using the dataset management tool FiftyOne~\cite{moore2020fiftyone}.

\begin{figure}
    \centering
    \includegraphics[width=\linewidth]{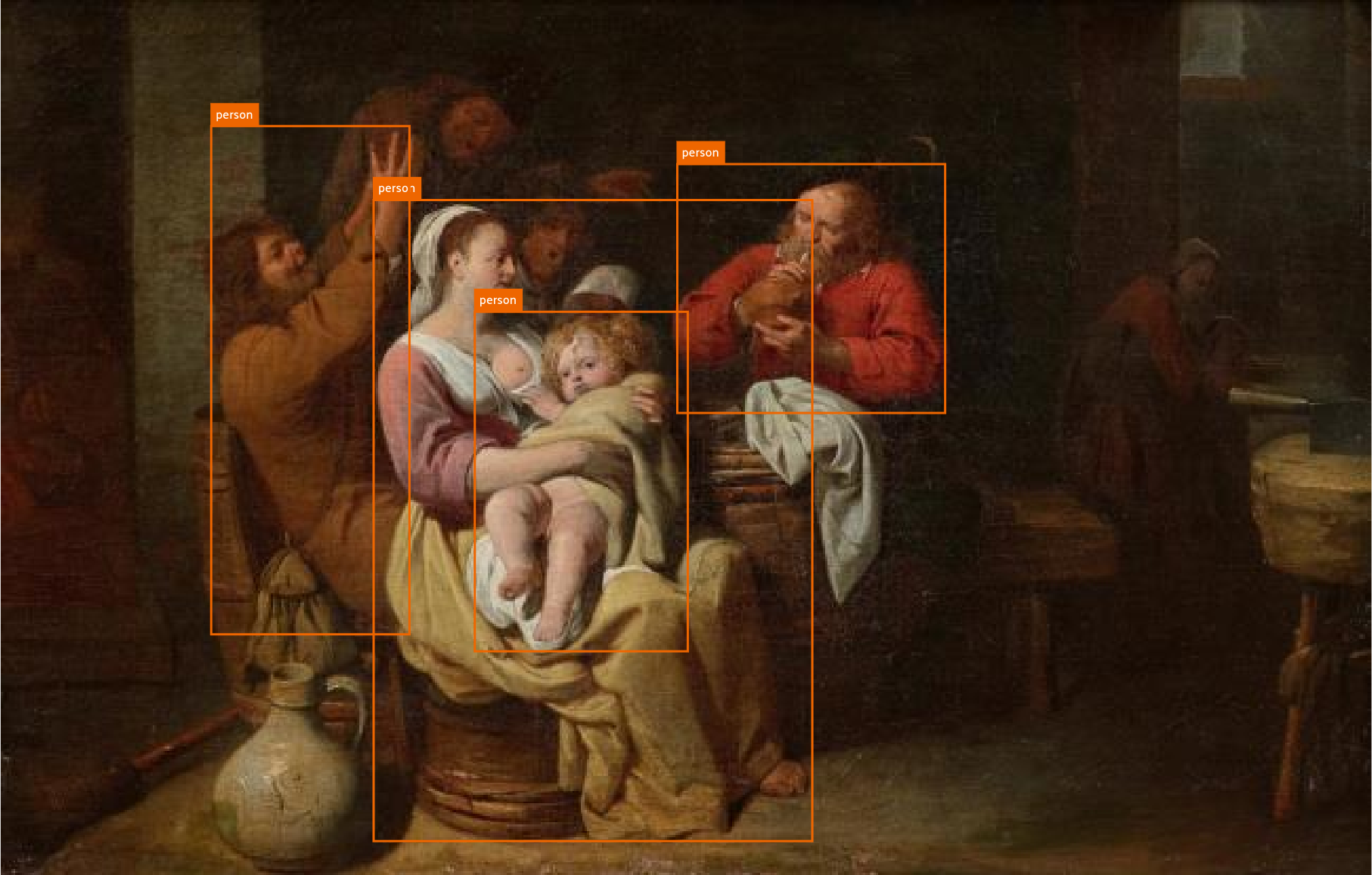}
    \caption{Example from the person annotation phase. While the four persons in the foreground were annotated with bounding boxes, the three persons in the background are hardly visible and were not annotated. Image credits:
    \textit{Company drinking and smoking in an interior}. David Rijckaert (III). 1627 -- 1661. Oil on canvas. RKD -- Netherlands Institute for Art History, RKDimages (301815). Public Domain.}
    \label{fig:box_ann}
\end{figure}

In the \textit{person annotation} phase, we annotated each person with tightly fitting bounding boxes and (possibly multiple) gesture labels.
Depending on the artwork style and reproduction quality, it can sometimes be difficult to distinguish between background and depicted persons. 
To handle these edge cases, we defined multiple requirements for image regions to be considered persons:
\begin{enumerate*}
    \item The head of the person must be visible.
    \item Apart from the head, at least two additional pose keypoints must be visible.
    \item It must be possible to assign this minimal set of keypoints to the person in question (in contrast to different, overlapping persons).
    \item It must be possible to clearly distinguish the persons from the background. %
\end{enumerate*}
\Cref{fig:box_ann} shows an example image where some of the depicted persons meet the criteria and are annotated and some others are not annotated.

These criteria, especially the third one, can be quite subjective.
There will always be instances where one annotator perceives a person as clearly distinguishable from the background, while another may not.
The person on the right corner of \cref{fig:box_ann} provides an example of such an edge case.
Given the diverse stylistic variations and artistic abstractions, we believe that encountering such edge cases is inevitable. 
We aim to address this issue by explicitly outlining the (unavoidably somewhat subjective) criteria in the annotation guidelines, yet we acknowledge that avoiding these ambiguities completely is not achievable.

\begin{figure}
    \centering
    \includegraphics[width=\linewidth]{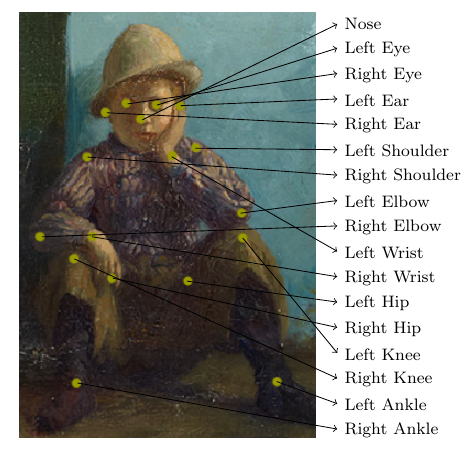}
    \caption{Illustration of Pose Estimation Keypoints. 
    Image Credits: Detail from \textit{Ein ruhiges Stündchen}. Ludwig Noster. 1895. Oil on Canvas. Alte Nationalgalerie, Staatliche Museen zu Berlin / Andreas Kilger. Public Domain.}
    \label{fig:example-kpts}
\end{figure}

\begin{figure}%
    \centering
    \includegraphics[width=\linewidth]{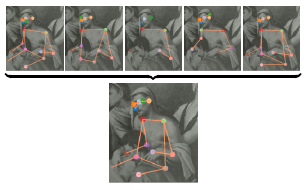}
    \caption{Example of an increase in annotation quality by merging multiple flawed annotations. Image credits:
    Detail from \textit{Die Auferweckung des Lazarus}. Bonifazio Veronese. ca. 1487 -- 1553. Deutsche Fotothek / Walter Möbius.}
    \label{fig:mergefig}
\end{figure}

Finally, in the \textit{keypoint annotation} phase, we applied the crowd-working platform AMT to annotate the cropped person boxes obtained in the second step with 17 keypoints.
Those points define the body pose as exemplified in \cref{fig:example-kpts}.
To ensure annotation quality, we gathered five annotations for each of the person boxes and merged them by averaging over the set of defined keypoint annotations. 
The annotation merging process can be defined as follows:

Let $\mathbf{k}^n_i = (x^n_i, y^n_i, v^n_i)$ denote the i-th keypoint annotated by the n-th annotator, where $(x^n_i, y^n_i, v^n_i)$ define the respective keypoint coordinates and visibility.
We encode the absence of the i-th keypoint annotation for the n-th annotator as $v^n_i = 0$ and define the set of present annotations indices $N = \{n_i | v^n_i \neq 0\}$.
The merged keypoint coordinates $\mathbf{k}^*_i$ are then given by:
\begin{equation}
    \mathbf{k}^*_i = (x^*_i, y^*_i)
\end{equation} with
\begin{equation}
    (x^*_i, y^*_i) =  \Big(\frac{1}{\lvert N \rvert} \sum_{n\in N} x^n_i,  \frac{1}{\lvert N \rvert} \sum_{n\in N} y^n_i \Big) 
    \;.
\end{equation}
In simpler terms, we construct the centroid of all annotated coordinates for each of the keypoints defining the body pose.
\Cref{fig:mergefig} illustrates how using this process multiple imperfect annotations can be merged into an accurate pose skeleton.

\subsection{Dataset Statistics}
The SniffyArt dataset consists of 1941 persons annotated with tightly fitting bounding boxes, 17 pose estimation keypoints, and gesture labels.
The annotations are spread over 441 historical artworks with diverse styles.
Note that the relatively low number of artworks is due to the difficulty of finding smell gestures in digital collections and we plan to extend the dataset in the future.
To the best of our knowledge, the current state of the SniffyArt dataset already constitutes the second-largest keypoint-level dataset in arts after the yet unpublished PoPArt~\cite{schneider2023poses}.
We provide predefined train, validation, and test splits, containing 307, 83, and 89 images, respectively (\cf \cref{tab:splits}) to facilitate training and enable a consistent baseline evaluation.
The splits were generated image-wise and based on the gesture labels, \ie person crops from the same image are always assigned to the same split, and the splits are used unmodified for all tasks.

\begin{table}[t]
    \caption{Overview of image, box, and gesture distribution for the dataset splits. Background class (\ie person not performing any smell gesture) is not listed. 
    Note that the splits were used unchanged for detection, keypoint estimation, and gesture classification and person boxes from one image were always assigned to the same split.
    }
    \centering
    \begin{tabular}{lccc}
    \toprule
         & Train & Validation & Test \\
         \midrule
        \# Images & \phantom{0}307 (\SI{64.1}{\percent}) & \phantom{0}83 (\SI{17.3}{\percent}) & \phantom{0}89 (\SI{18.5}{\percent}) \\
        \# Persons & 1245 (\SI{64.1}{\percent}) & 332 (\SI{17.1}{\percent}) & 364 (\SI{18.8}{\percent}) \\
        \# Gestures & \phantom{0}434 (\SI{60.4}{\percent}) & 127 (\SI{17.7}{\percent}) & 130 (\SI{18.1}{\percent}) \\
        \bottomrule
    \end{tabular}
    \label{tab:splits}
\end{table}

\begin{figure}
    \centering
    \includegraphics[width=.8\linewidth]{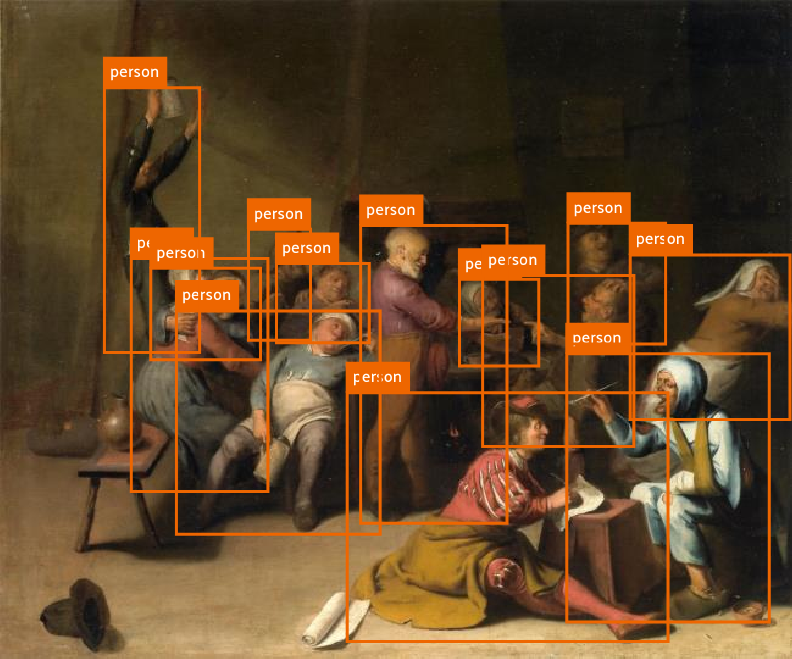}
    \caption{Example of a large group of persons where only three out of twelve depicted persons perform a smell gesture.
    Image credits: \textit{Carousing peasant company in an inn}. Joachim van den Heuvel. Oil on panel. RKD -- Netherlands Institute for Art History,  RKDimages (284006). Public Domain.}
    \label{fig:background_persons}
\end{figure}

Due to our choice to annotate all persons meeting the requirements defined in \cref{sec:coll} irrespective of whether they perform a smell gesture, we observe a large class imbalance with background persons (\ie performing no smell gesture) being vastly overrepresented (\cf \cref{fig:distb}.
\Cref{fig:background_persons} shows an example from the dataset where only three of the twelve annotated persons perform a smell gesture while the remaining nine are labelled as background persons.
While the resulting imbalance negatively affects the performance of gesture classification, it was necessary to enable complete annotations for person and keypoint detection algorithms.
However, without considering the background class, the class imbalance is reduced considerably as illustrated in \cref{fig:dista}.
\begin{figure}
\begin{subfigure}{\columnwidth}
    \subcaption{Excluding background class.}
    \label{fig:dista}
    \centering
    \begin{tikzpicture}
        \begin{axis} [
            xmin=0,
            xmax=6,
            ybar,
            ymin=0,
            ymax=170,
            xtick={1,...,5},
            bar width=.3cm,
            xticklabels={smoking, drinking, holding the nose, cooking, sniffing},
            xmin=0.4,
            xmax=5.7,
            x tick label style={rotate=45, anchor=east},
            ymajorgrids,
            axis lines=left,
            ylabel=Number of Instances,
            axis line style={-}
            ]
            \addplot coordinates {
            (1,160)
            (2,123)
            (3,101)
            (4,32)
            (5,31)
            };
            \addplot coordinates {
             (1,45)
             (2,43)
             (3,21)
             (4,16)
             (5,13)
            };
            \addplot coordinates {
             (1,52)
             (2,45)
             (3,24)
             (4,10)
             (5,12)
            };
            \legend {Train, Valid, Test};
        \end{axis} 
    \end{tikzpicture}
\end{subfigure}
\begin{subfigure}{\linewidth}
    \subcaption{Including background class and multi-labels. Note that multi-class labels (smoking, drinking) are not mutually exclusive with their single-class constituents, \ie a person annotated as smoking and drinking is counted three times in this distribution.}
    \label{fig:distb}
\centering
        \begin{tabular}{lccc}
    \toprule
        Gesture & Train & Valid & Test  \\
        \midrule
        Background & 795 & 198 & 230 \\
        Smoking & 160 & \phantom{0}43 & \phantom{0}52 \\
        Drinking & 123 & \phantom{0}45 & \phantom{0}45 \\
        Holding the Nose & 101 & \phantom{0}21 & \phantom{0}24 \\
        Cooking & \phantom{0}32 & \phantom{0}16 & \phantom{0}10 \\
        Sniffing & \phantom{0}31 & \phantom{0}13 & \phantom{0}12  \\
        Smoking, Drinking & \phantom{0}13 & \phantom{0}11 & \phantom{0}10 \\
        \bottomrule 
    \end{tabular}
\end{subfigure}
    \caption{Distribution of gesture class labels in train and test set. }
    \label{fig:classdist}
\end{figure}

We allowed persons to be annotated with multiple gestures, effectively rendering the classification problem as multi-label classification. 
In practice, we encountered more than thirty examples of persons smoking and drinking at the same time (\cf \cref{fig:distb}) but no other combinations.
However, for future extensions of the dataset, different label combinations are to be expected.

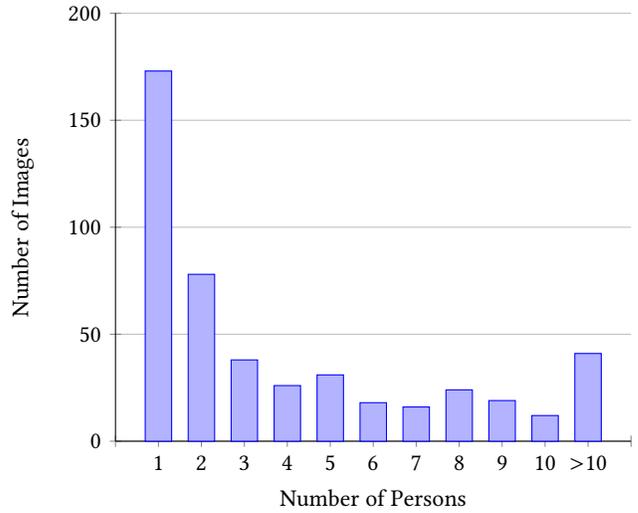
\begin{figure}
    \centering
    \begin{tikzpicture}
        \begin{axis} [
            xmax=12,
            xmin=0,
            ybar,
            ymin=0,
            ymax=200,
            xtick={0,...,12},
            xticklabels={,1,...,10,$>$10},
            ymajorgrids,
            ylabel = Number of Images,
            xlabel = Number of Persons,
            axis lines=left,
            axis line style={-}
            ]
            \addplot coordinates {
            (1,173)
            (2,78)
            (3,38)
            (4,26)
            (5,31)
            (6,18)
            (7,16)
            (8,24)
            (9,19)
            (10,12)
            (11,41)
            };
        \end{axis} 
    \end{tikzpicture}
    \caption{Distribution of annotated persons per image.}
    \label{fig:box_dist}
\end{figure}

The distribution of the number of depicted persons per image (\cf \cref{fig:box_dist}) reflects the remarks about the high number of background persons.
While  \SI{53}{\percent}  of the images contain only one or two persons, a considerable amount of images depict 10 or more persons. 

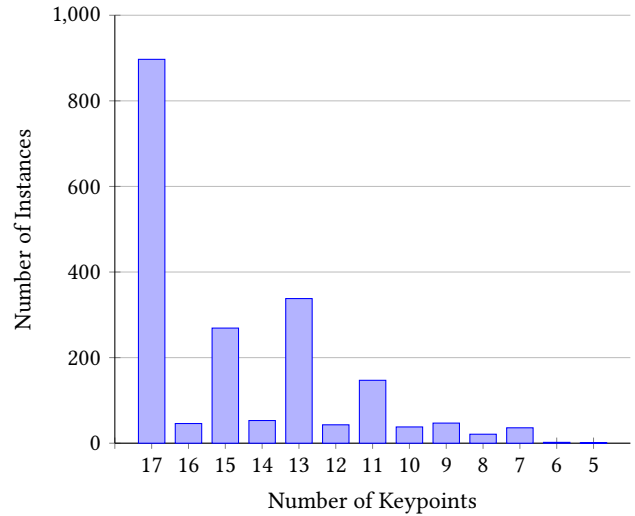
\begin{figure}
    \centering
    \begin{tikzpicture}
        \begin{axis} [
            xmax=18,
            xmin=4,
            ybar,
            ymin=0,
            ymax=1000,
            xtick={18,...,5},
            xticklabels={,17,...,5},
            ymajorgrids,
            x dir = reverse,
            ylabel = Number of Instances,
            xlabel = Number of Keypoints,
            axis lines=left,
            axis line style={-}
            ]
            \addplot coordinates {
            (17,897)
            (16,46)
            (15,269)
            (14,53)
            (13,338)
            (12,43)
            (11,147)
            (10,38)
            (9,47)
            (8,21)
            (7,36)
            (6,2)
            (5,1)
            };
        \end{axis} 
    \end{tikzpicture}
    \caption{Distribution of annotated keypoints per person.}
    \label{fig:kpt_dist}
\end{figure}

Regarding the distribution of annotated keypoints per person we observe that the majority of person boxes \SI{46}{\percent} have annotations for each of the 17 possible keypoints, while only \SI{6}{\percent} have annotations for less than 10 keypoints (\cf \cref{fig:kpt_dist}).

\subsection{Distribution Format}
The annotations are provided in a JSON file following the COCO standard for object detection and keypoint annotations. 
Extending the default COCO format, we enrich each entry in the annotations array of the COCO JSON with a \say{gestures} key that contains a (possibly empty) list of smell gestures the annotated person is performing.
To facilitate label transformation for single-label classification, we add a derived \say{gesture} key, which contains the list of gesture labels as a single, comma-separated string.
Additionally, we provide a CSV file with image-level metadata, which includes content related-fields such as Iconclass codes or image descriptions, as well as formal annotations, such as artist, license or creation year.
For license compliance, we do not publish the images directly.
Instead, we provide links to their source collections in the metadata file and a Python script to download the artwork images.
The dataset is available for download on Zenodo.\footnote{\url{https://doi.org/10.5281/zenodo.8273616}}

\section{Baseline Experiments}
To showcase the applicability of our dataset and provide initial baselines, we conduct experiments for person detection, human pose estimation, and gesture classification. 
\href{spreadsheet}{}
\subsection{Detection}
Detecting depicted persons is a prerequisite for both, top-down approaches in keypoint estimation, and person-level gesture classification. 
Here, we evaluate three representative object detection configurations:
\begin{enumerate*}
    \item Faster R-CNN~\cite{ren2015faster} with a ResNet-50~\cite{he2016deep} serves as a default baseline as it is still the most widely used object detection system.
    \item To assess the effect of scaling up the backbone, we evaluate a Faster R-CNN with the larger ResNet-101~\cite{he2016deep} backbone.
    \item To understand the effects of more modern detection heads, we evaluate the state-of-the-art transformer-based DINO~\cite{zhang2022dino} architecture with a ResNet-50 backbone.
\end{enumerate*}

All models are trained for 50 epochs using the MMDetection~\cite{mmdetection} framework, applying the respective default training parameters. 
Please refer to \cref{tab:det_params} for a detailed list of hyperparameters.
\begin{table*}[t]

    \centering
    \caption{Fine-tuning settings of detection and pose estimation experiments.}
    \begin{tabular}{lllll}
    \toprule
        \thead[l]{Parameter} & \thead{Faster R-CNN \\ RN-50/RN-101} & \thead{DINO \\ RN-50} & \thead{Pose HRNet \\ HRNet-W32} & \thead{DEKR \\ HRNet-W32}\\
         \midrule
         task & detection & detection & pose estimation & pose estimation \\
         \midrule
         pre-training dataset & ImageNet-1k & ImageNet-1k & ImageNet-1k & ImageNet-1k  \\
         optimizer & SGD & AdamW & Adam & Adam \\
         base lr & 0.02 & 0.0001 & 0.001 & 0.0005\\
         weight decay & 0.0001 & 0.0001 & - & -\\
         optim. momentum & 0.9 & - & - & - \\
         batch size & 2 & 2 & 10 & 64 \\
         num\_gpus & 2 & 2 &2 &2\\
         training epochs. & 50 & 50 & 210 & 210\\
         warmup iterations & 500 & - & 500 & 500 \\
         warmup scheduler & linear & - & linear & linear \\
         lr scheduler & step (30,40,48) & step (11) & step (170, 200) & step (170, 200)\\
         lr gamma & 0.1 & 0.1 & 0.1 & 0.1\\
         \bottomrule
    \end{tabular}
    \label{tab:det_params}
\end{table*}

In \cref{tab:detection_results}, we report the model performances, following the standard COCO evaluation protocol\footnote{\url{https://cocodataset.org/\#detection-eval}} for object detection. 
For each configuration, we fine-tune five models on the SniffyArt training set and report their respective mean and standard deviation of test set performance. 

\begin{table*}[t]
    \centering
    \caption{COCO detection performance of representative detection algorithms fine-tuned on SniffyArt-train and evaluated on SniffyArt-test, averaged over five runs. The standard deviation is reported in brackets.}
    \begin{tabular}{lccccccc}
    \toprule
         Model & Backbone & $AP$ & $AP_{50}$ & $AP_{75}$ & $AP_{M}$ & $AP_{L}$ & $AR$ \\
         \midrule 
         Faster R-CNN~\cite{ren2015faster} & ResNet-50~\cite{he2016deep} & 34.2($\pm$0.02) & 75.5($\pm$0.10) & 24.6($\pm$0.17) & 22.6($\pm$0.24) & 35.9($\pm$0.05) & 43.0($\pm$0.04) \\
         Faster R-CNN~\cite{ren2015faster} & ResNet-101~\cite{he2016deep} &  35.7($\pm$0.08)  &  76.0($\pm$0.13)  &  28.5($\pm$0.21)  &  24.3($\pm$0.10) &  37.4($\pm$0.09)  &  44.0($\pm$0.08)  \\ 
         DINO~\cite{zhang2022dino} & ResNet-50~\cite{he2016deep} &  28.4($\pm$0.09)  &  53.7($\pm$0.31)  &  27.0($\pm$0.09)  &  15.4($\pm$0.16)  &  30.1($\pm$0.15)  &  61.7($\pm$0.04) \\
         \bottomrule
    \end{tabular}
    \label{tab:detection_results}
\end{table*}

Despite the relatively small size of the dataset, we observe an increase of \SI{1.5}{\percent} mAP in detection accuracy when scaling up the feature extraction backbone.
While the configuration equipped with a ResNet-101 outperforms its ResNet-50 counterpart in the stricter $AP_{75}$ metric considerably (\SI{+3.9}{\percent}), the difference in $AP_{50}$ amounts to only \SI{0.5}{\percent}. 

This suggests that the larger backbone mostly increases the model's capacity to localize persons very precisely. %
Surprisingly, the modern DINO architecture performs considerably worse (\SI{-5.8}{\percent}) than its Faster R-CNN counterpart.
Noticeable is the high recall of the DINO models, which is \SI{18}{\percent} higher than that of the Faster R-CNN counterpart with the same backbone.
We hypothesize that the DINO models generate too many box predictions for the images in our datasets and that the performance can significantly be increased by filtering out weak predictions or reducing the number of object queries. 

We conclude that standard architectures with relatively weak backbones can already produce sufficient person predictions based on the size of the SniffyArt training set. 
If required, more accurate boxes can likely be obtained by pre-training using external data (\eg DeArt~\cite{reshetnikov2022deart}, COCO~\cite{lin2014microsoft}, or \cite{westlake2016detecting}) or scaling up model capacity.

\subsection{Pose Estimation}
To understand how different human pose estimation paradigms work for our dataset, we analyse one top-down method (DEKR) and one bottom-up method (Pose HRNet). 
In the top-down scenario, the pose estimation model gets the box predictions from our strongest detection model as an auxiliary input at validation and test time.
We use the MMPose~\cite{mmpose2020} framework for model training, initialize the backbones with ImageNet-1k weights and train for 210 epochs using the default hyperparameters. 
For more details on training settings, please refer to the two rightmost columns of \cref{tab:det_params}.
Again, we fine-tune five models on the SniffyArt training set and report the mean and standard deviation on the SniffyArt test set in \cref{tab:hpe_results}.
As the evaluation metric, we apply COCO's object keypoint similarity (OKS)\footnote{\url{https://cocodataset.org/\#keypoint-eval}}.
\begin{table*}[]
    \centering
    \caption{Performance of representative human pose estimation (HPE) algorithms fine-tuned on SniffyArt-train and evaluated on SniffyArt-test, averaged over five runs. The standard deviation is reported in brackets.}
    \begin{tabular}{lccccccc}
    \toprule
         Model & Backbone &  $AP$ & $AP_{50}$ & $AP_{75}$ & $AP_{M}$ & $AP_{L}$ & $AR$ \\
         \midrule 
         Pose HRNet~\cite{sun2019deep} & HRNet-W32~\cite{wang2020deep} &  53.3($\pm$0.05)  &  79.2($\pm$0.07)  &  58.3($\pm$0.17)  &  30.7($\pm$0.13)  &  56.3($\pm$0.07)  &  58.8($\pm$0.05)  \\
         DEKR~\cite{geng2021bottom} & HRNet-W32~\cite{wang2020deep} &   36.7($\pm$0.10)  &  70.0($\pm$0.08)  &  35.6($\pm$0.19)  &  13.7($\pm$0.07)  &  43.4($\pm$0.07)  &  45.8($\pm$0.06) \\
         \bottomrule
    \end{tabular}
    \label{tab:hpe_results}
\end{table*}

The results show that while keeping the backbone invariant, the bottom-up pipeline DEKR is significantly outperformed by the top-down approach Pose HRNet in all metrics.
\subsection{Gesture Classification}
We analyze the performance of various representative networks for the classification of smell gestures. 
Experiments are conducted per-person, meaning that each person is cropped and classified separately. 
To simplify our models, we transform the multi-label problem into a single-label classification by introducing new labels representing combinations of single labels. %
Effectively, this required the introduction of only a single new class, since drinking and smoking is the only combination of smell gestures present in the dataset.
We apply cross-entropy loss and handle the class imbalance by weighing it with normalised inverse class frequencies. %
Experiments are conducted using the MMPretrain~\cite{2023mmpretrain} framework keeping the default parameters for the classification algorithms.
As for detection and keypoint estimation, we fine-tune five models and report the average top 1 accuracy, precision, and $F_1$ scores together with the standard deviations in \cref{tab:cls_results}. 
Additionally, we report the metrics of a naive classifier that always predicts the majority class.
    
\begin{table}
    \centering
    \caption{Classification results on the SniffyArt test set for three classification networks pre-trained on ImageNet-1k. \emph{Majority Class} denotes the trivial solution of always predicting the most frequent class (\ie \emph{no gesture}). We report the mean over five experiments per configuration with standard deviation in brackets.}
    \label{tab:cls_results}
    \begin{tabular}{lccc}
    \toprule
       Model &  Acc./top1 & Prec. & $F_1$\\
        \midrule
    Majority Class &  9.1  &  14.3  &  11.1  \\
       ResNet-50~\cite{he2016deep} &
        31.8($\pm$3.7)  &  31.8($\pm$1.4)  &  31.1($\pm$2.0) \\
       ResNet-101~\cite{he2016deep} &
        36.7($\pm$3.8)  &  34.1($\pm$1.2)  &  34.2($\pm$1.8)  \\
       HRNet-W32~\cite{wang2020deep} &
        15.7($\pm$1.5)  &  19.8($\pm$2.1)  &  17.3($\pm$1.8)  \\
       \bottomrule
    \end{tabular} 
\end{table}
The evaluation highlights how challenging the classification of odor gestures on historical artworks is. 
While we do see an increase in the metrics when increasing the number of model parameters, the overall $F_1$ score stays quite low with \SI{34}{\percent}.
Surprisingly, the performance of the modern HRNet falls significantly behind that of the two ResNet models.
We note that this performance gap is consistent over the evaluations of all trained models which is reflected in the relatively low standard deviations in all metrics.

To assess how well feature representations learned from person detection and keypoint estimation generalise to the gesture classification task, we initialize the networks with weights obtained from the feature extraction backbones of the person detection and keypoint estimation tasks discussed above. 
With similar experimental settings, we train five models for each configuration and report the results in \cref{tab:transfer}.
\begin{table}
    \centering
    \caption{Classification performance when initializing the feature extraction backends with weights obtained by person detection (for ResNet-50 \& ResNet-101), or keypoint estimation (for HRNet-W32) on the SniffyArt dataset.}
    \label{tab:transfer}
    \begin{tabular}{lccc}
    \toprule
       Model &  Acc./top1 & Prec. & $F_1$\\
        \midrule
       ResNet-50~\cite{he2016deep} &
        12.6($\pm$1.2)  &  16.5($\pm$0.7)  &  14.1($\pm$0.9)  \\
       ResNet-101~\cite{he2016deep} &
        15.0($\pm$2.0)  &  18.4($\pm$2.2)  &  16.1($\pm$2.0)  \\
       HRNet-W32~\cite{wang2020deep} & 
        37.7($\pm$4.1)  &  33.3($\pm$2.3)  &  33.7($\pm$2.8)  \\
       \bottomrule
    \end{tabular} 
\end{table}
When comparing the ResNets pre-trained for person detection with their ImageNet-pretrained counterparts, we observe a significant performance drop, with $F_1$ scores decreasing by over half.
This suggests that feature representations learned from person detection are not suited for smell gesture classification.
The HRNet models, on the other hand, seem to benefit greatly from initializing them with weights obtained by keypoint estimation. 
We find that the weak performance metrics of the ImageNet-pretrained models are more than doubled when keypoint estimation pre-training is used.
This finding demonstrates the large potential of combining the representational space from the two tasks of gesture classification and keypoint estimation.

\section{Limitations}
\paragraph{Dataset Size}
With 400 images, the number of annotated artworks is relatively low. 
This is due to the difficulties in finding a sufficient number of smell gestures in artworks which can partly be explained by a lack of olfaction-related metadata in digital museum collections~\cite{ehrich2022nose}. 
We plan to extend the dataset in the future, alleviating this issue by applying semi-automated approaches based on the set of existing images.
\paragraph{Annotation Quality}
During the test runs of the keypoint annotation phase, we observed that annotators often incorrectly left out occluded keypoints, even if they were inside of the image boundaries.
To alleviate this problem, we incorporated pose keypoints, even if they were annotated by only one of the five annotators. 
However, this approach may lead to incorrect annotations if one annotator misunderstands the task or provided incorrect keypoints deliberately. 
To prevent such cases, a more advanced outlier detection algorithm could be implemented to filter out annotations from obstructive annotators.
\paragraph{Experimental Evidence}
To confirm and strengthen the hypothesis that leveraging pose estimation keypoints is beneficial for smell gesture classification, more experiments would be needed.
A deeper analysis is out of the scope of this paper but it would certainly be a valuable line of future research to investigate the combination potential further. 
\paragraph{Image Properties}
The degree of artistic abstraction and low quality of some of the images might set an upper bound to algorithmic gesture recognition capabilities.
While extensions with regard to dataset size and the incorporation of different digital collections might alleviate this issue to some degree, it is a general problem of computer vision algorithms in the artistic domain.
From another angle, it might as well be viewed as a strength as it enforces algorithm robustness towards diverse stylistic representations.
\section{Conclusion}
We introduced the SniffyArt dataset consisting of 1941 persons on 441 historical artworks, annotated with tightly fitting bounding boxes, 17 pose estimation keypoints and gesture labels.
By combining detection, pose estimation, and gesture labels, we pave the way for innovative classification approaches connecting these annotations.
Our dataset features high-quality human pose estimation keypoints, which are achieved through merging five distinct sets of keypoint annotations per person.
In addition, we have conducted a compre\-hen\-sive baseline analysis to evaluate the performance of various representative algorithms for detection, keypoint estimation, and classification tasks. 
Preliminary experiments demonstrate that there is a large potential in combining keypoint estimation and smell gesture classi\-fication tasks.
Looking ahead, we plan to extend the dataset and address the relatively low number of samples.
Given the scarcity of metadata related to olfactory dimensions in digital museum collections, we intent to apply semi-automated approaches to identify candidate images containing smell gestures. 
Even in its current state, the SniffyArt dataset provides a solid foundation for the development of novel algorithms focused on smell gesture classification. 
We are particularly interested in exploring multi-task approaches that leverage both pose keypoints and person boxes.
As we move forward, we envision that this dataset will stimulate significant advancements in the field, ultimately enhancing our understanding of human gestures and olfactory dimensions in historical artworks.

\section{Acknowledgements}
This paper has received funding from the Odeuropa EU H2020 project under grant agreement No.\ 101004469. 
We gratefully acknowledge the donation of the NVIDIA corporation of two Quadro RTX 8000 that we used for the experiments.

\clearpage
\bibliographystyle{ACM-Reference-Format}
\balance
\bibliography{refs}

\end{document}